\documentclass[10pt,twocolumn,letterpaper]{article}

\usepackage{wacv}
\usepackage{times}
\usepackage{epsfig}
\usepackage{graphicx}
\usepackage{amsmath}
\usepackage{amssymb}
\usepackage{subcaption}
\usepackage{multirow}
\usepackage{booktabs}
\usepackage{breqn}
\usepackage{amsmath}
\usepackage[square,sort,numbers]{natbib}
\usepackage{stfloats} % for image positioning

\usepackage{xcolor}
\definecolor{green_fig_legend}{RGB}{0, 255, 0}
\definecolor{magenta_fig_legend}{RGB}{255, 0, 255}

\newcommand{\todo}[1]{}
\renewcommand{\todo}[1]{{\color{red} TODO: {#1}}}

\wacvfinalcopy % *** Uncomment this line for the final submission

\def\wacvPaperID{694} % *** Enter the wacv Paper ID here

\ifwacvfinal\fi
\begin{document}

%%%%%%%%% TITLE
\title{Print Defect Mapping with Semantic Segmentation}

% Authors at the same institution
%\author{First Author \hspace{2cm} Second Author \\
\author{Augusto C.\ Valente \hspace{1cm} Cristina Wada \hspace{1cm} Deangela Neves \hspace{1cm} Deangeli Neves\\
F\'abio V.\ M.\ Perez \hspace{1cm} Guilherme A.\ S.\ Megeto \hspace{1cm} Marcos H.\ Cascone \hspace{1cm} Otavio Gomes \\
Instituto de Pesquisas Eldorado\\
%{\tt\small \{augusto.valente, cristina.wada, deangela.neves, deangeli.neves,\\
%\tt\small fabio.perez, guilherme.megeto, marcos.cascone, otavio.gomes\} @eldorado.org.br}
\tt\small firstname.lastname@eldorado.org.br
\and
Qian Lin \\
HP Labs, HP Inc.\\
{\tt\small qian.lin@hp.com}
}

\maketitle
\ifwacvfinal\thispagestyle{empty}\fi

%%%%%%%%% ABSTRACT
\begin{abstract}
Efficient automated print defect mapping is valuable to the printing industry since such defects directly influence customer-perceived printer quality and manually mapping them is cost-ineffective. Conventional methods consist of complicated and hand-crafted feature engineering techniques, usually targeting only one type of defect.
In this paper, we propose the first end-to-end framework to map print defects at pixel level, adopting an approach based on semantic segmentation. Our framework uses Convolutional Neural Networks, specifically DeepLab-v3+, and achieves promising results in the identification of defects in printed images. We use synthetic training data by simulating two types of print defects and a print-scan effect with image processing and computer graphic techniques.
Compared with conventional methods, our framework is versatile, allowing two inference strategies, one being near real-time and providing coarser results, and the other focusing on offline processing with more fine-grained detection. Our model is evaluated on a dataset of real printed images.
\end{abstract}

%%%%%%%%% BODY TEXT
\section{Introduction}

The absence of print defects is one of the most significant factors that influence customer satisfaction when experiencing a printer product. For this reason, mapping such defects before the final revision in preparation for market launch is extremely important to the manufacturing printer industry.

In large-scale quality control processes, print defect detection still requires human experts in an approach that is repetitive, time-consuming, and prone to human error and subjective evaluation~\cite{wang2016}. In order to improve the efficacy and efficiency of that process, automated methods to map print defects are highly desired.

There is a wide variety of print defects that may arise in the printing process, such as streaks, banding, spots, and patches. The causes of each of them vary according to the printer technology type. In this paper, we mainly focus on banding and streaks, since they are easier to be synthesized, and are both common to inkjet and laser electrophotographic printers.

In inkjet printers, banding is generally caused by small electromechanical or chemical problems in the printer components, such as damaged nozzles in the print heads, or ink density variation~\cite{briggs2000}. Streak defects, on the other hand, can be a result of misaligned or contaminated print heads.

In laser printers, the non-uniform line spacing observed in banding defects is mainly caused by fluctuations in the velocity of the optical photoconductor (OPC) drum~\cite{jang2005simulation}. Moreover, imperfections in the cleaning mechanism on the intermediate transfer belt (ITB), damages in the belt surface itself, or the color cartridges may generate streak artifacts on printings.

Some works have proposed automated intelligent systems for assessing print quality~\cite{Verikas:2011:RAC:1994471.1994864}. Conventionally adopted approaches~\cite{wang2016, spivakovsky2018image, xiao2017, yangping2018real} focus on print quality defect detection using standard machine vision methods, coarsely indicating the location of print defects. However, none of these solutions employ defect detection at the pixel level and map different types of defects in a single model. In this work, we specifically attack this problem, which we refer to as \textit{print defect mapping}.

Since 2012, deep convolutional neural networks (DCNNs) have achieved remarkable results in many computer vision tasks, including image classification, object detection, and semantic segmentation~\cite{long2015fcns,ronneberger2015unet,chen2018deeplabv2}. In this paper, we employ semantic segmentation to detect and map print artifacts at the pixel level. We use DeepLab-v3+~\cite{chen2018deeplabv3plus} as the backbone of our framework. We chose this particular architecture because of its potential to simultaneously retrieve rich contextual information and sharp object boundaries, which are essential for characterizing print defects.

While deep learning techniques show impressive results in multiple computer vision tasks, they usually require large annotated datasets. Manually collecting and annotating data can be both expensive and time-consuming. When it comes to acquiring well-annotated data for training semantic segmentation models, which requires fine-grained label maps, that task becomes even more difficult and costly~\cite{zlateski2018importance}.

To overcome that problem, we create artificial print artifacts samples with image processing techniques to generate our training dataset. We initially collect a set of digital images with diverse content and add the dark streaks and the color bandings. In addition, we simulate the effect of printing and scanning documents by estimating color transformations from real devices. With that, we aim to make our training data close to real use cases.

Inspired by works in image quality assessment~\cite{bosse2018, kang2014}, we also implement two different paradigms for print defect mapping: No-Reference (NR) and Full-Reference (FR).
In the first one, the framework receives only the printed images, not requiring the original printable image. In the second paradigm, both the printed and the original digital images form the input, which may give additional cues to where the defects may occur.

Our main contributions are as follows:

\begin{itemize}
    \item  We present an end-to-end deep learning framework for pixel-wise print defect mapping. To the best of our knowledge, this is the first work to map print defects at the pixel level.

    \item  We propose a pipeline for simulating a print-and-scan effect and to artificially create print artifacts in diverse background images for two types of defects---dark streaks and banding. The proposed pipeline also leverages automatic pixel annotation, resulting in cost-free data in terms of human annotation.

    \item Our framework supports two input modes, Full-Reference (FR) and No-Reference (NR), that can be interchangeably used with small changes to its architecture. Moreover, it also adopts two inference strategies: \textit{Resized image} and \textit{Patch-based}, taking a trade-off between speed and quality into account.

\end{itemize}

%-------------------------------------------------------------------------
\section{Literature Review}\label{sec:overview}

\subsection{Print Defect Mapping}\label{subsec:pdm}

Previous approaches to Print Defect Mapping focus only on the coarse detection of defects. Generally, no more than one defect type is targeted, eliminating the need for mapping each input to a category. Furthermore, the detection task usually relies on the application of a set of standard image operations and similarity measures.

Wang \etal~\cite{wang2016} investigated the detection of local print defects in the form of spots and patches by analyzing the standard deviation of pre-extracted regions of interest. For each region, they use the valley-emphasis threshold to find potential local defects. Then, a set of hand-crafted features, such as edge strength and average RGB value, are extracted and fed into an SVM model to determine whether the overall quality of the printed imaged is good enough. They evaluate their model in terms of accuracy for global quality prediction, instead of localized defect detection as we propose.

Zhang \etal~\cite{runzhe2019} adopted a similar approach to detect light streaks in printer test images. Firstly, they pre-process standard print test pages in order to select only the smooth areas of the image. They divide these areas into blocks and compute the projection of the distance between each pixel within a block and the average pixel value in the block to detect streaks. Finally, they use logistic regression to remove false-positives from the previous step.

Spivakovsky \etal~\cite{spivakovsky2018image} proposed a SSIM-based comparison between printable images and their scanned versions to detect defects. Similarly, Xiao \etal~\cite{xiao2017} aligned reference and scanned images with feature-based registration, and compared them based on color difference distribution to detect text fading. Yangping \etal~\cite{yangping2018real} proposed a system based on grayscale and gradient differences for template matching.

While standard image processing techniques may work when detecting a specific type of artifact, it is challenging to cover multiple occurrences of different print defect types in the same image. Unlike conventional methods, our framework takes full advantage of deep neural networks, which have proved to generalize remarkably well in several computer vision tasks, to map both dark streaks and color banding defects pixel-wise.

\subsection{Semantic Segmentation}

Semantic segmentation is a computer vision task that consists of assigning a label to every pixel in an image.
Results on public benchmarks have considerably improved with the use of fully-convolutional DCNNs, such as FCN~\cite{long2015fcns}, U-Net~\cite{ronneberger2015unet}, and DeepLab~\cite{chen2018deeplabv2}. Applications of semantic segmentation are diverse, including medical image analysis~\cite{ronneberger2015unet}, remote sensing~\cite{remote_sensing_xx_zhu} and autonomous driving~\cite{siam2018rtseg}. In this work, we also employ semantic segmentation techniques, specifically DeepLab-v3+, to the problem of print defect mapping.

The motivation behind the application of semantic segmentation to print defect mapping is the precision gains in locating and identifying the defective patterns using a more detailed inspection. Besides, this approach would provide better generalization since the model will focus on deviations of the ideal printer operation rather than on specific characteristics of the defect.

\subsection{Print Defect Simulation}

Obtaining real data to train a print defect mapping algorithm is hard for several reasons. One would first need to use multiple defective printers and print several documents in order to obtain a good range of defects and backgrounds. Additionally, carefully annotating defect locations in the images would take a long time. As deep learning methods require significant amounts of labeled data, we generate synthetic training data using image processing techniques. That consisted in simulating the actual print defects and also the overall look of a printed-and-scanned page. We detail the implementation in the next section.

We based our artifact model in \cite{jang2005simulation}, where the authors created a framework to reproduce defects from laser printers. They estimate parameters from actual print defects and use both geometrical modeling and scanned templates. However, their approach still depends on printing the pages to achieve realism.

\section{Methodology}\label{sec:methodology}

\subsection{Synthetic data}\label{subsec:artifacts}

The ability to create realistic synthetic artifacts has many advantages. First, it can generate large amounts of training data, which is a demand for deep neural networks. Second, it can produce different types of artifacts, such as fine pitch banding and streaks, that generally occurs uniformly on the printed page; or repetitive artifacts, like spots, which may occur randomly across the page.

\begin{figure}[ht]
     \centering
     \begin{subfigure}[b]{0.4\columnwidth}
         \centering
         \includegraphics[width=\textwidth]{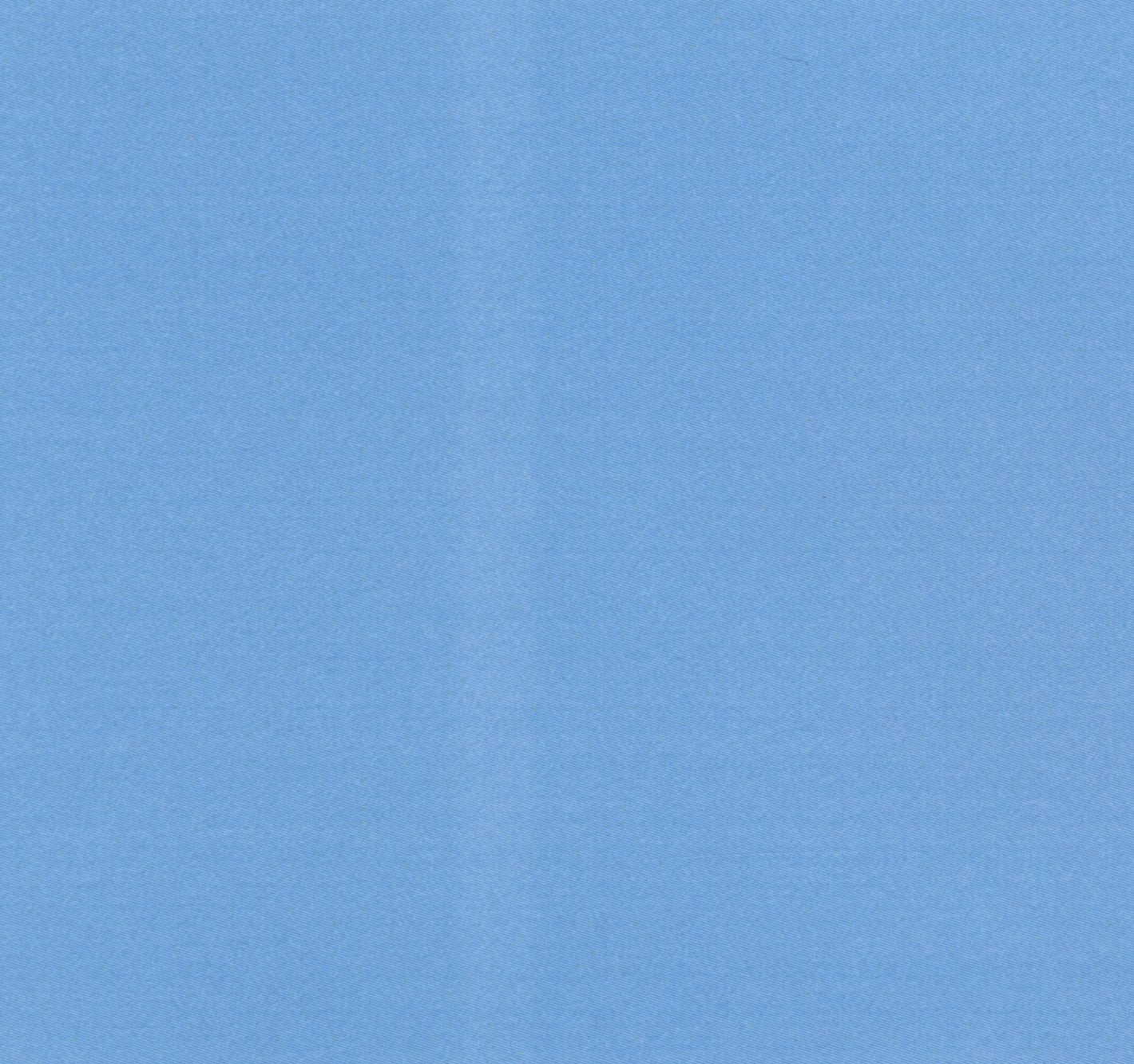}
         \caption{}
         \label{fig:banding_example}
     \end{subfigure}
     \begin{subfigure}[b]{0.4\columnwidth}
         \centering
         \includegraphics[width=\textwidth]{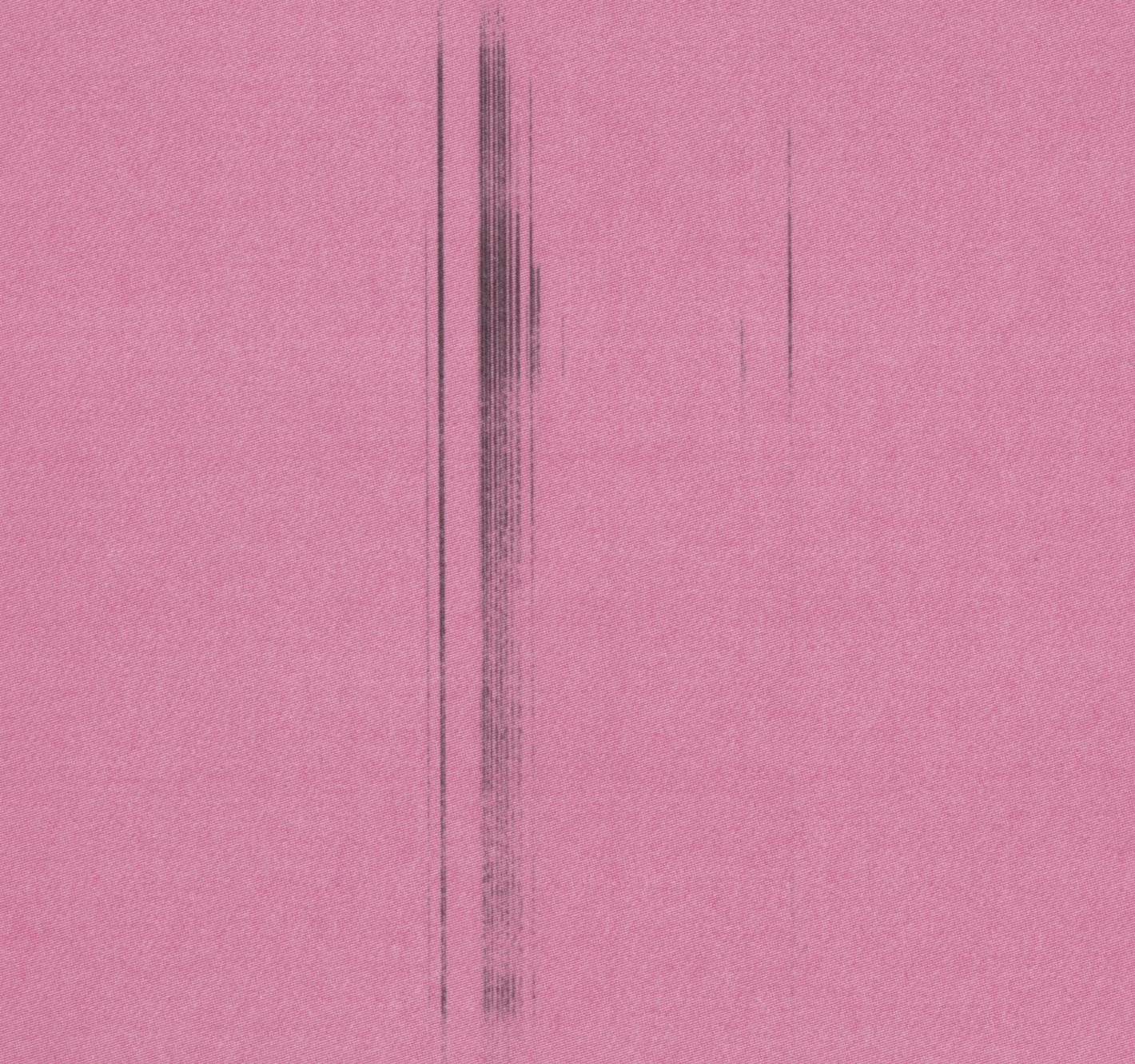}
         \caption{}
         \label{fig:streak_example}
     \end{subfigure}
        \caption{Print defect examples: (a) color banding with an increase in the magenta channel, and (b) dark streak.}
        \label{fig:artifacts_real_examples}
\end{figure}

We address two frequent types of print defects: dark streaks and color bandings. They essentially result in horizontal or vertical lines with an excess or lack of a certain color component. Dark streaks are usually thinner textured lines, while color bandings are rectangular regions with near homogeneous drop or excess in a color channel. Figure~\ref{fig:artifacts_real_examples} shows examples of such defects.

\begin{figure*}[ht]
    \centering
    \begin{subfigure}{.22\linewidth}
        \centering
        \includegraphics[height=0.12\textheight]{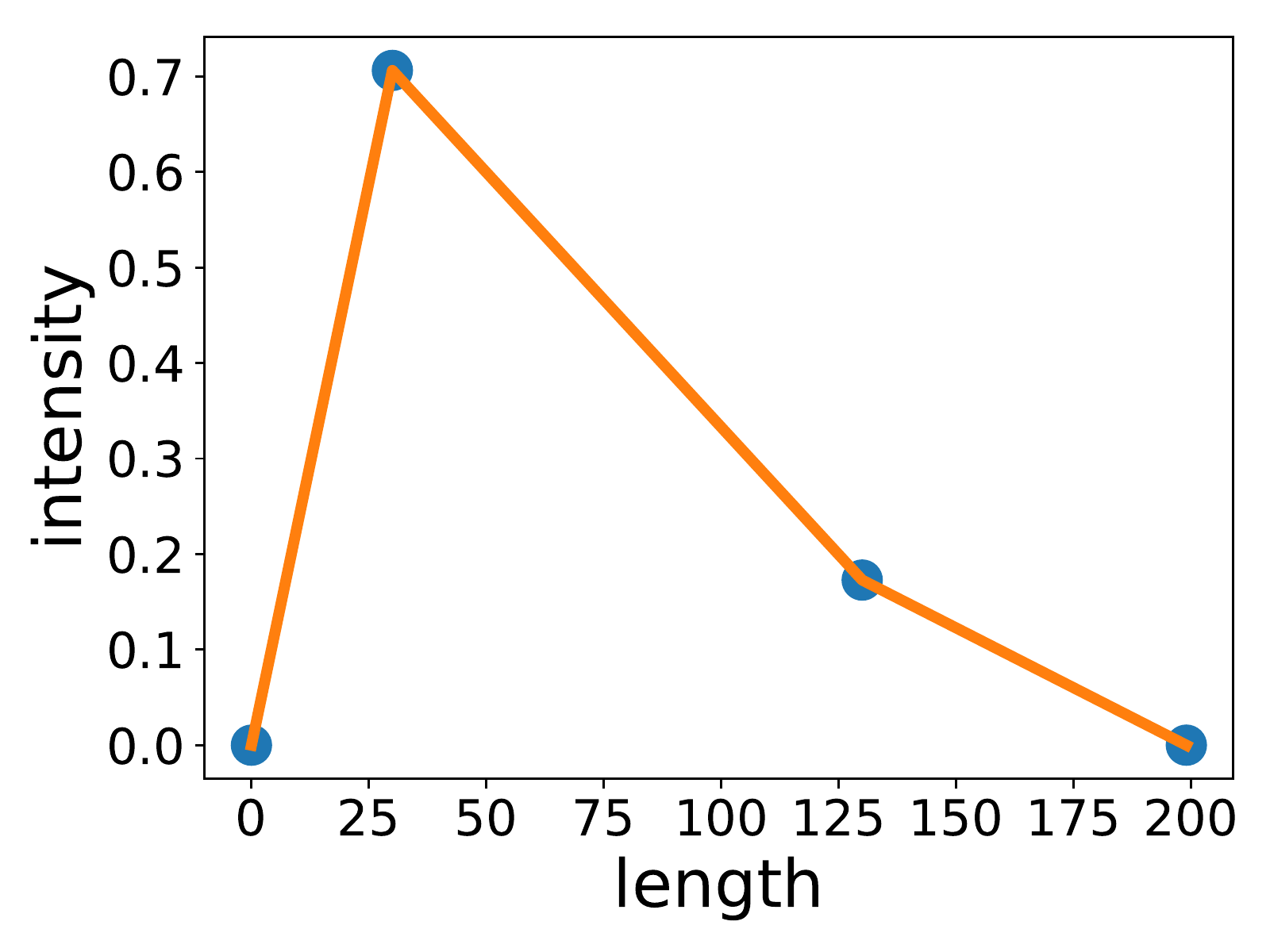}
        \caption{}
    \end{subfigure} \vspace{-0.1cm}
    \begin{subfigure}{.22\linewidth}
        \centering
        \includegraphics[height=0.12\textheight]{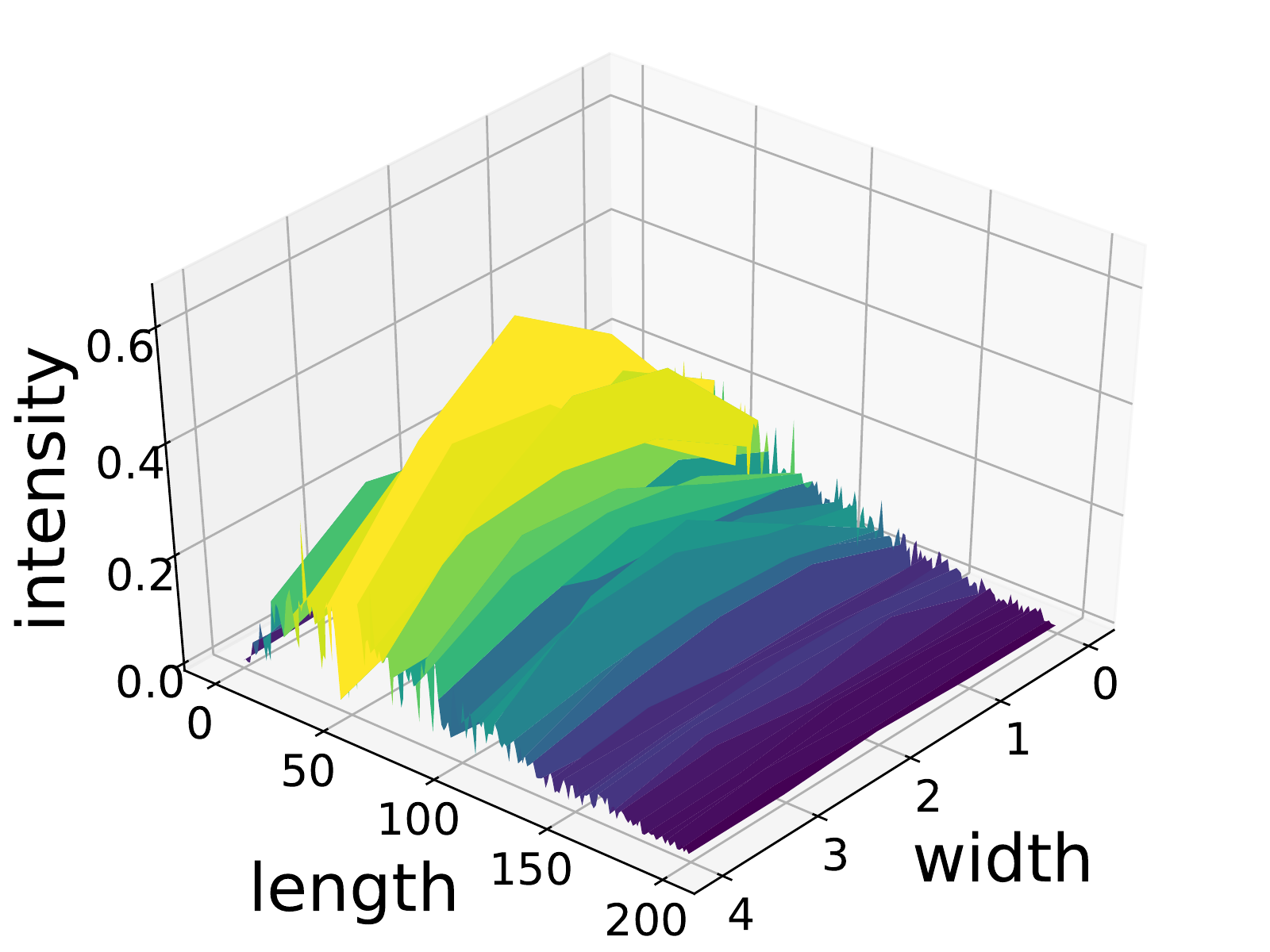}
        \caption{}
    \end{subfigure}
        \begin{subfigure}{.08\linewidth}
        \centering
        \includegraphics[height=0.12\textheight]{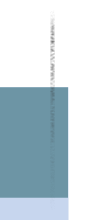}
        \caption{}
    \end{subfigure} 
    \begin{subfigure}{.18\linewidth}
        \centering
        \includegraphics[height=0.12\textheight]{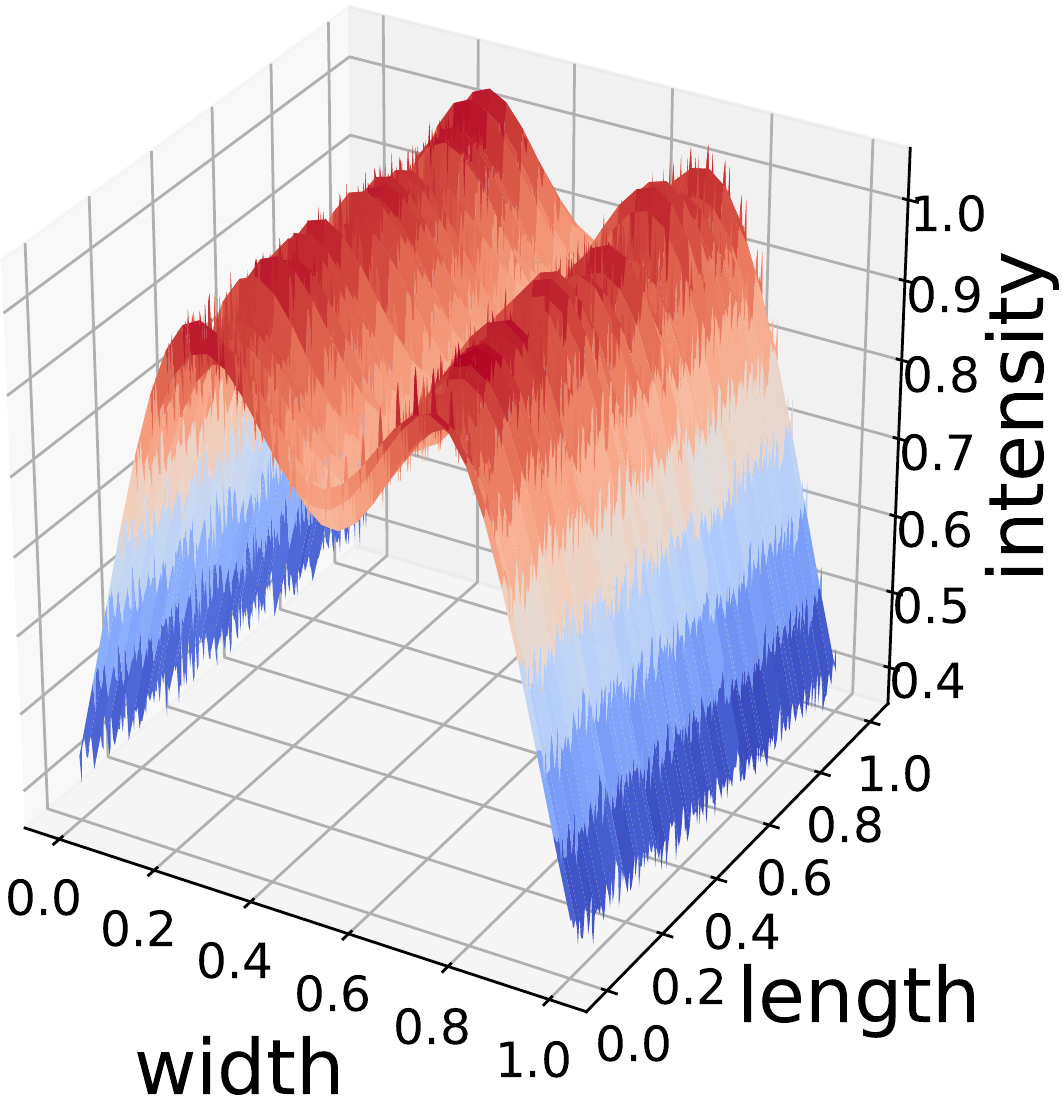}
        \caption{}
    \end{subfigure} 
    \begin{subfigure}{.1\linewidth}
        \centering
        \includegraphics[height=0.12\textheight]{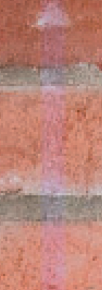}
        \caption{}
    \end{subfigure}
    \caption{Examples of pixel intensity profile of dark streak and color banding simulations. Dark streak: (a) intensity profile across the defect length, (b) 3D intensity profile and (c) the corresponding simulated defect. Color banding: (d) 3D intensity profile and (e) color banding defect applied to the magenta channel. Best viewed on screen.}
    \label{fig:simulations}
\end{figure*}

To simulate dark streaks, we blend a random thin rectangular image region with a dark-colored texture pattern. This texture pattern was procedurally generated by applying Perlin noise to create a gray level perturbation. The pattern varies the color intensity, randomly along the streak length, and the edges are set to be lighter than the center. Figure~\ref{fig:simulations} shows an example of a simulated dark streak.

When simulating color banding, on the other hand, we first convert the input image to CMYK, which is the color space adopted in the printing process. After that, we overlay rectangular regions that affect each color channel separately and have slight variations in color. Pixel intensities can be either decremented/incremented to simulate lack/excess. The amount of pixel intensity modification varies across the band's width, being either brighter or darker in the middle. This variation is subject to a bimodal Gaussian distribution, as shown in Figure~\ref{fig:simulations}.

The last step in our synthetic data pipeline is to imitate the appearance of printing and scanning an image. This process consists of learning the RGB tone curves of printer and scanner devices. We printed and scanned test pages in different devices and modeled the color transformation as a polynomial, whose coefficients are determined by applying a linear least square (LLS) regression between the original digital image and its respective printed-and-scanned version on a real device.

\vspace{-0.5cm}
\begin{multline} \label{eqn:1}
F(x,y, R,G,B) = a_0 + a_{1}y + a_{2}x + a_{3}R + a_{4}G + a_{5}B \\ +a_{6}yR + a_{7}yG + a_{8}yB + a_{9}xR + a_{10}xG + \\ a_{11}xB 
+a_{12}RG + a_{13}RB + a_{14}GB + a_{15}RGB
\end{multline}

The input variables of the polynomial function shown in Equation \ref{eqn:1} are, respectively, the normalized image x and y coordinates, and its respective red, green, and blue color channels. Figure \ref{fig:simulations} shows an example result of the print-scan simulation process.

\begin{figure}[ht]
  \centering
    \begin{subfigure}{.45\linewidth}
        \centering
        \includegraphics[width=\textwidth]{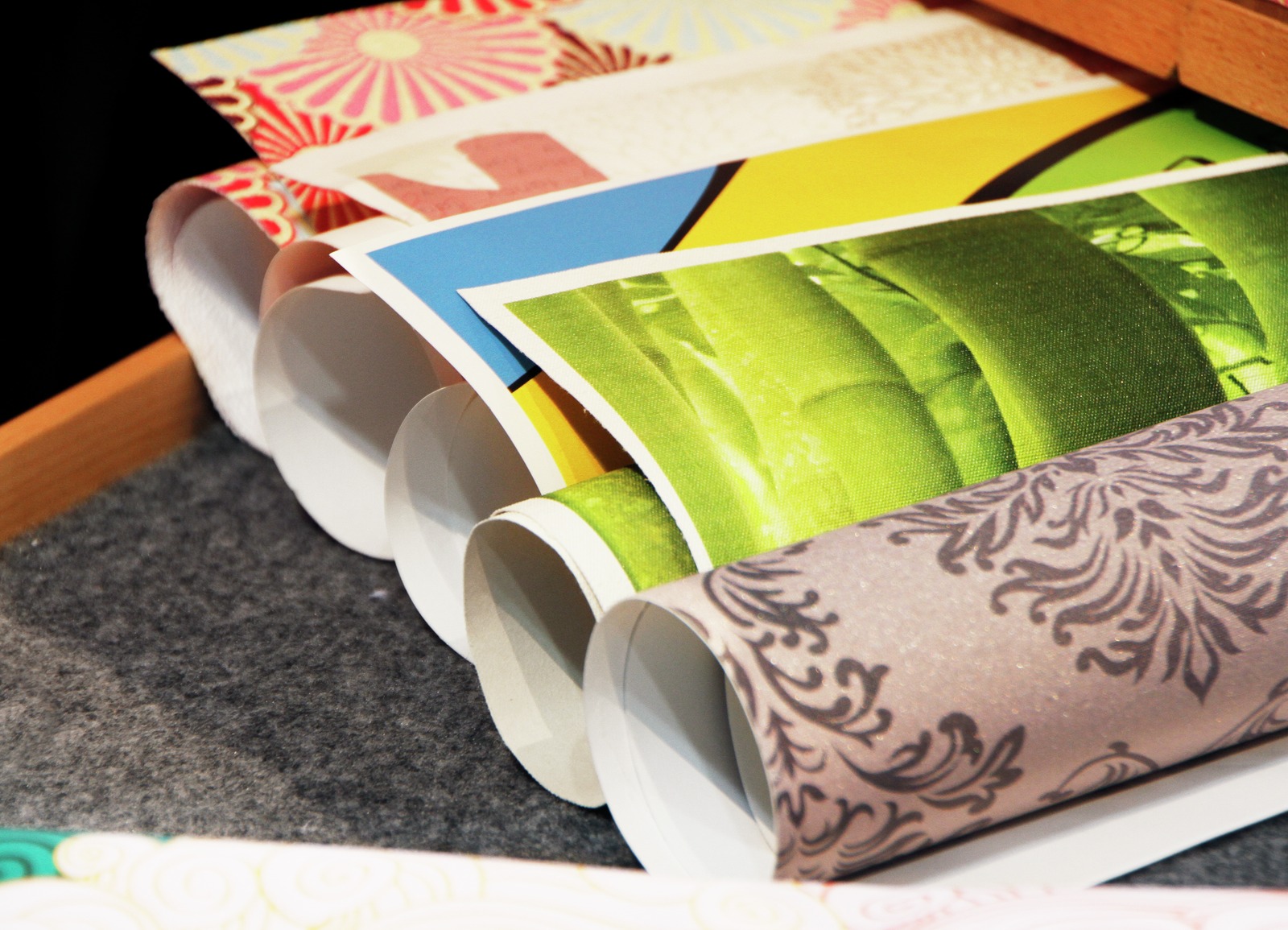}
        \caption{}
    \end{subfigure}
    \begin{subfigure}{.45\linewidth}
        \centering
        \includegraphics[width=\textwidth]{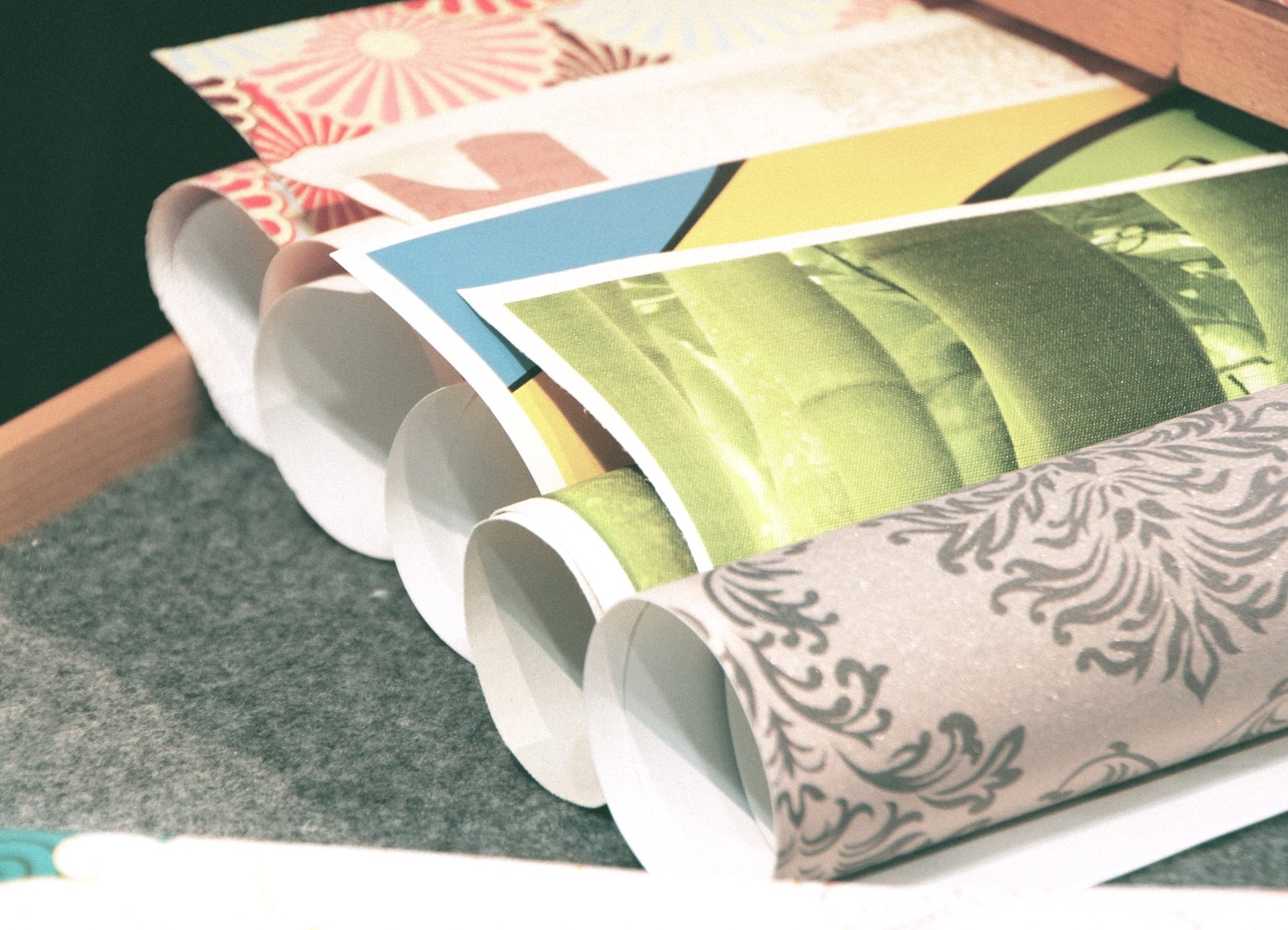}
        \caption{}
    \end{subfigure}
    \caption{Print-scan simulation example: (a) original image and (b) result after the effect is applied.}
    \label{fig:print_scan_simulation}
\end{figure}

%--------------------------------------------------

\subsection{Network architecture}

As mentioned in Section~\ref{sec:overview}, considering the potential of print defect mapping using a semantic segmentation approach, we investigated some DCNN architectures, particularly U-Net~\cite{ronneberger2015unet} and DeepLab-v3+~\cite{chen2018deeplabv3plus}, since both attain state-of-art results in other tasks. In our preliminary experiments, DeepLab-v3+ outperformed U-Net when segmenting thinner lines. For this reason, we decided to use DeepLab-v3+ architecture as a base for our model.

DeepLab-v3+~\cite{chen2018deeplabv3plus} combines the benefits of two methods: an atrous spatial pyramid pooling (ASPP) module \cite{chen2018deeplabv2} and an encode-decoder structure. 
In encoding-decoding configurations, the network constructs a feature representation of the image in lower resolution and then applies trained fractionally-strided convolutions to recover label maps in higher resolution. The low-resolution intermediary representation provides good context understanding but hinders the detection of small or thin objects. In order to compensate for this issue, the ASPP module, based on dilated or atrous convolutions, leverages its increased receptive field to avoid sacrificing spatial resolution, and recovers labeled pixels by applying bilinear upsampling. Less use of downsampling helps fine-grained detection, but the context understanding is not as broad as in encoder-decoder networks. Another advantage of using DeepLab-v3+ architecture is that it does not require huge datasets, as it was trained on the Pascal VOC 2012 dataset \cite{chen2018deeplabv3plus}.

To configure DeepLab-v3+ for print defect mapping, we performed variations in the architecture hyperparameters and evaluated their impact on the defect identification.

The ASPP module performs multiple pooling scales through four parallel atrous convolutions using a set of 256 filters of size $1\times1$ or $3\times3$ in dilated configuration and using depth-wise separable convolutions. The encoder-decoder structure, in its turn, has a DCNN in the encoder backbone, performing atrous convolution followed by the ASPP structure. In the decoder module, the DCNN output is concatenated with the bilinearly upsampled ASPP output. The concatenated result is convolved with a $3\times3$ filter and bilinearly upsampled to the desired output. 

We tested different output strides (8 and 16), a parameter that controls the ratio of input image resolution to the output image resolution. In general, an output stride of 8 gave better results, but at the cost of slower training. In the DCNN, we verified the use of ImageNet-pretrained ResNet-50 and ResNet-101 models. The former achieved similar performance while being faster to train.

ResNet-50 is composed of building blocks, each one formed by stacked sets of 3 convolutional layers. We tested different dilation values---$[1, 1, 1]$, $[1, 2, 1]$, and $[1, 2, 4]$---in these 3 layers, achieving the best performances with $[1, 2, 1]$, and $[1, 1, 1]$ in some cases.

\begin{figure*}[b]
    \begin{subfigure}{.48\textwidth}
      \centering
        \includegraphics[width=.95\linewidth]{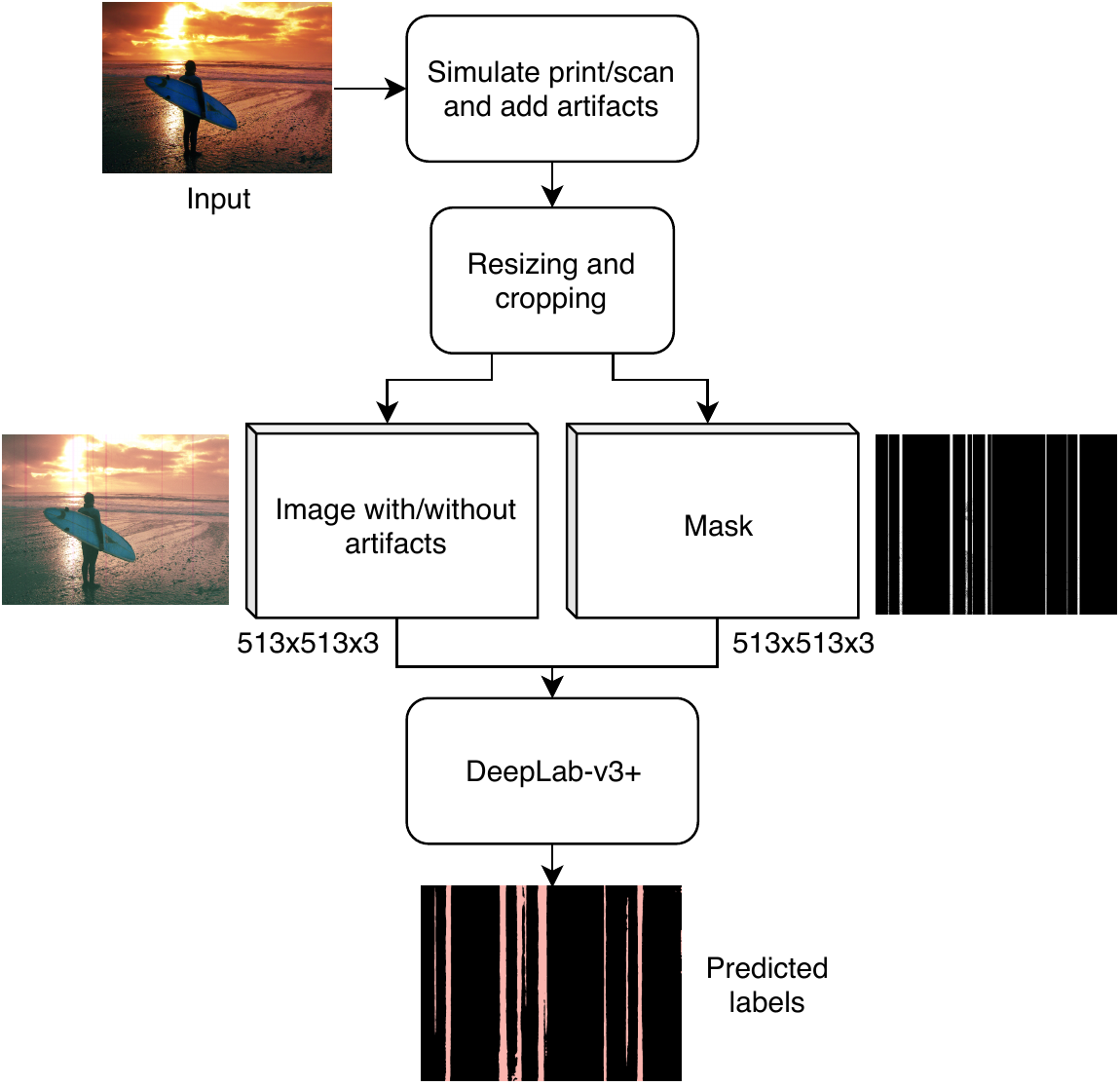}
        \caption{}
        \label{fig:nr-pipeline}
    \end{subfigure}
    \begin{subfigure}{.48\textwidth}
      \centering
        \includegraphics[width=.95\linewidth]{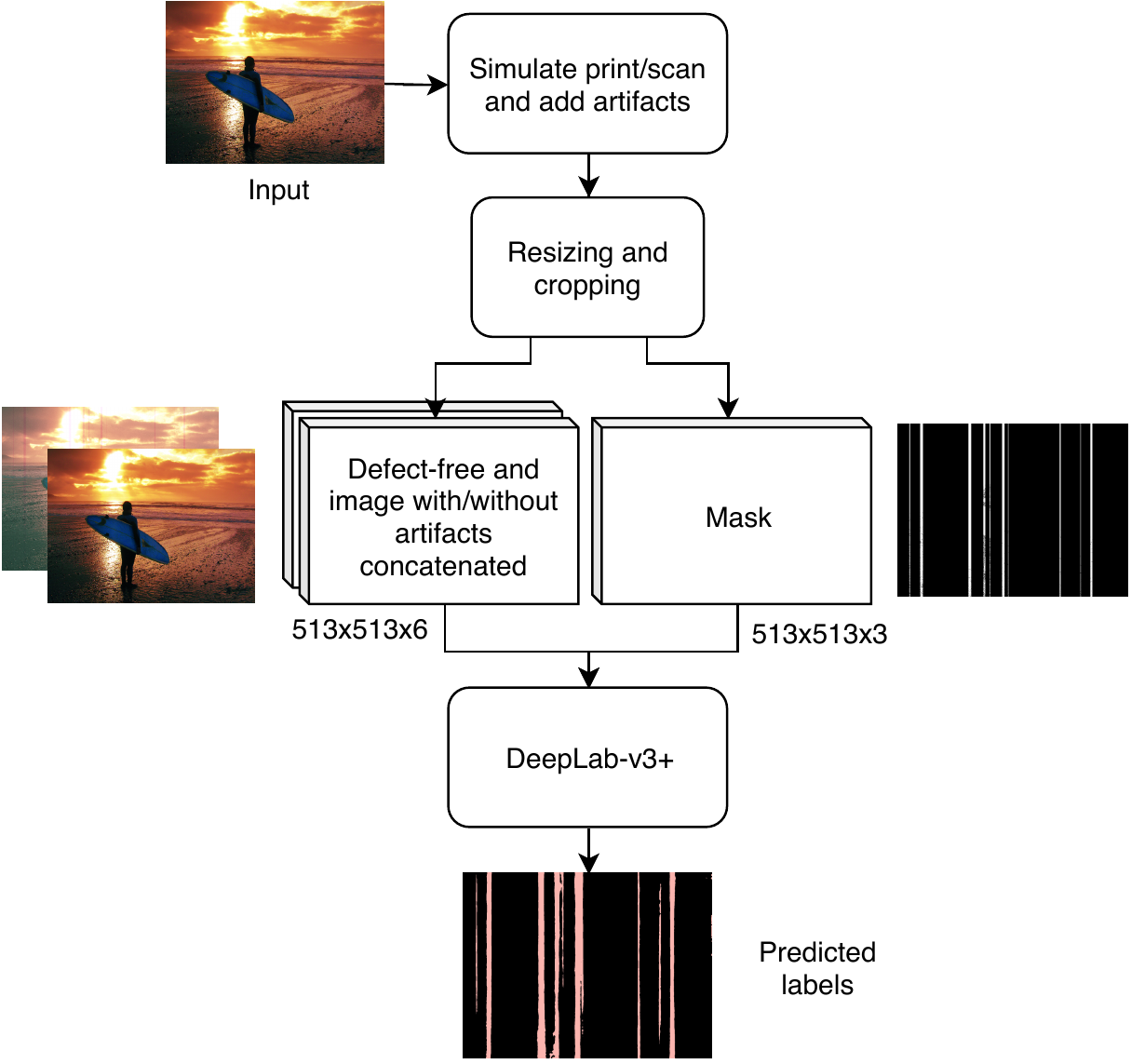}
        \caption{}
        \label{fig:fr-pipeline}
    \end{subfigure}
    \caption{Training pipeline for (a) no-reference and (b) full-reference methods.}
    \label{fig:short}
\end{figure*}

\begin{figure*}[ht]
    \centering
    \begin{subfigure}{0.7\textwidth}
        \centering
        \includegraphics[width=\textwidth]{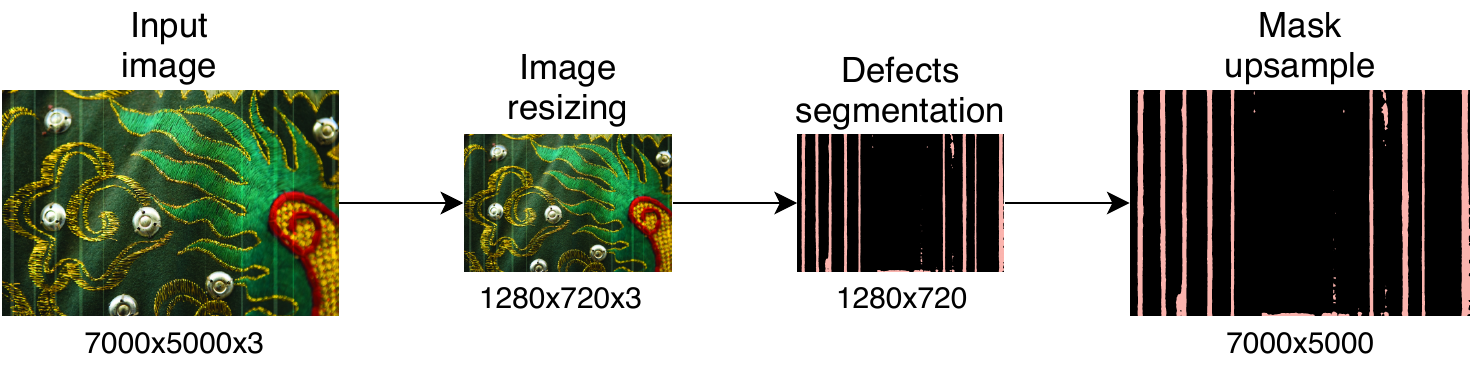}
        %\caption{}
    \end{subfigure} \\ \vspace{-0.1cm}
    \begin{subfigure}{0.7\textwidth}
        \centering
        \includegraphics[width=\textwidth]{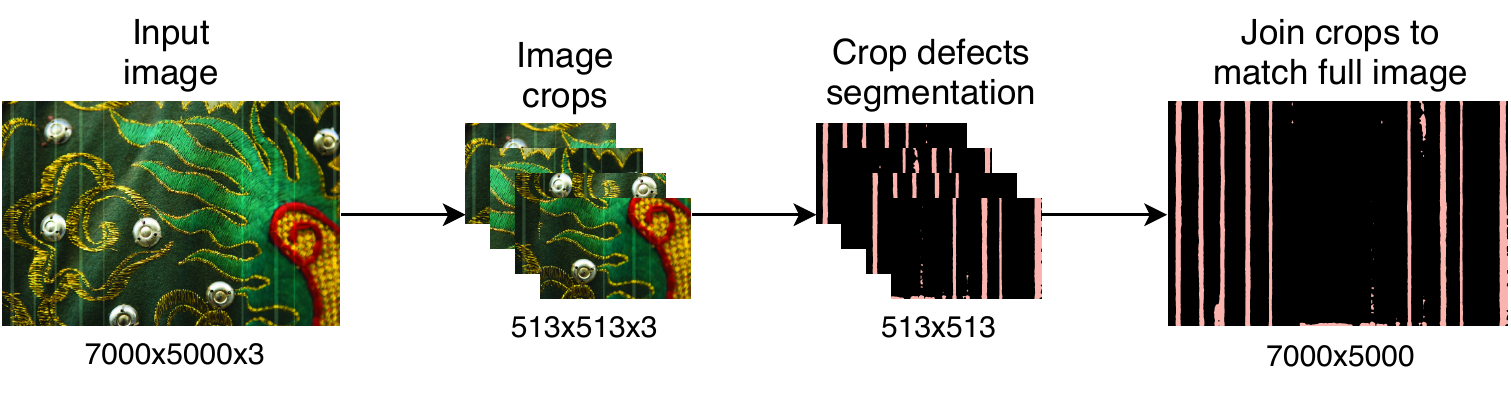}
        %\caption{}
    \end{subfigure} 
    \vspace{-0.1cm}
    \caption{Inference pipeline with resized image approach (top) and patch-based approach (bottom).}
    \label{fig:inference}
\end{figure*}

\subsection{No-Reference and Full-Reference metrics}\label{subsec:iqa}

We designed the framework of this work to support two metrics used in Image Quality Assessment (IQA). 

IQA is a broad research field that focuses on quantitative representing the human perception of image quality. IQA is generally divided into two areas: Full-Reference (FR) and No-Reference (NR). FR explores metrics that quantify the quality of a distorted image, given its reference image. On the other hand, NR metrics measure the quality of distorted images without requiring their reference image.

FR methods are usually based on image operations. The existence of reference images provides useful information about their distorted counterparts, resulting in a wide range of evaluation possibilities. In contrast to that, having no ground truth image to establish quality criteria makes NR a more challenging problem, which has seen more recent advances with the rise of deep learning~\cite{li2016_nr_iqa_dnn,kim2017fully_deep_blind,bosse2018,pan2018blind_predicting}.

Both FR and NR extend their application to other fields in which measuring output image quality is essential to improve the end-user viewing experience, such as Image Capturing and Image Processing. Since print defects are directly linked to the overall quality of printed images, we added both sets of metrics.

\subsection{Generating training data}

Our training set consists of $2\,949$ digital images---including documents, photos, and slides---with minimum dimensions of $1920\times1080$. The samples are preprocessed online during training by a set of techniques, described in Section \ref{subsec:artifacts}, that generate their potentially defective printed versions. Thus, the virtual number of training samples is much higher than the raw number of digital images in the dataset, since each image results in potentially infinite defective versions of itself.  % resulting in approximately $589\,800$ training samples.

We first simulate the appearance of printing and scanning effects on digital images. Then, we randomly decide if the image will be defective based on an assigned probability of 0.9. If so, we randomly add the artifacts described in Section~\ref{subsec:artifacts} to it. The probability of streaks occurrence varied from $0.3$ to $0.5$, while the sum of channel banding probabilities is its complementary. Finally, we generate the ground truth mask indicating which pixels have artifacts. 

We resized all processed images and their respective masks to a common size of $1920 \times 1080$ pixels using bicubic interpolation, and keeping their aspect ratio. Their respective masks are interpolated with nearest neighbors to preserve label information.

\subsection{Training pipeline} \label{subsec:train_strategy}

To train the NR network, we extract random patches of $513\times513$ pixels of the processed image and its respective mask. Then, we feed this set into the network, which considers the corresponding set of masks as the ground truth. Figure~\ref{fig:nr-pipeline} shows a diagram of the full training pipeline for the NR network.

For FR, we extract random patches of $513\times513$ pixels of the reference/defective images and the mask. We concatenate the extracted patches from reference and defective images forming a 6-channel image, which is fed to the network, as shown in Figure ~\ref{fig:fr-pipeline}. The change to the network input introduces more weights in the first layer. To initialize them, we simply copied the pretrained weights from the first three channels, since the new channels also correspond to an RGB image. We also tried to use the difference between the reference and target images as the network input. However, we found that concatenation provided better results.

We used the multi-class cross-entropy loss and weighted pixels labeled as background with $0.05$ to account for the imbalance between background and defects. We explored the intersection over union loss~\cite{berman2018lovasz} but did not get better results. We also tried to train a one-vs-all class setting, implemented as the sum of the binary cross-entropy loss for each class against all other classes but got worse results.

We used the Adam optimizer \cite{KingmaB14} with an initial learning rate of $10^{-4}$, and weight decay of $10^{-5}$. We adopted a strategy of accumulating gradients between batches. This allows us to achieve a larger effective batch size. We used a batch size of 5, and accumulate gradients for 10 iterations, resulting in an effective batch size of 50.

We employed online data augmentation during training (90-degree clockwise and counter-clockwise rotation and horizontal flipping). We normalized each image by subtracting its mean and dividing by its standard deviation. We randomly split the dataset into $90\%$ for training and $10\%$ for validation. We also verified if we increased the number of training samples to $11\,102$ would give better results, but no significant improvement was observed.

%Since the DCNN models demands large amounts of training data, We also verified if the increasing the number of training samples to $11\,102$ would give better results, but no significant improvement was observed.

\subsection{Inference strategy} \label{subsec:inf_strategy}

We evaluate our framework considering two approaches: \textit{resized image} and \textit{patch-based}. In the first, we resize images down to a resolution of $1280\times720$ with bicubic interpolation before feeding it to the model. In the second, we extract patches of $513\times513$ pixels from the original image and run inference separately on each of them. In the end, we combine the predictions from each image patch.

These inference strategies are supported by both FR and NR pipelines and are illustrated in Figure~\ref{fig:inference}.

\section{Experiments}\label{sec:experiments}

% Experiment IDs
% FR
%  - 20191002-brj451gvg7-2
%  - 20191007-4155cbh-2
%  - 20191007-brj451gvg9-6
%  - 20191007-hp-z840-2
% NR
%  - 20190930-brj451gvg9-2
%  - 20191002-brj451gvg9-1
%  - 20191002-4155cbh-2
%  - 20190919-brj451gvg9-1
\begin{table}[b]
\begin{center}
\caption{Evaluation on semi-synthetic and real data for both inference strategies and full-reference (FR) and no-reference (NR) methods. Values indicate the average and standard deviation for four runs, to account for randomness.}
\vspace{-0.15cm}
\label{tab:scores}
\begin{tabular}{llrrr}
\toprule
\textbf{Method} & \textbf{} & \multicolumn{1}{c}{\textbf{mIoU}} & 
\multicolumn{1}{r}{\textbf{\begin{tabular}[c]{@{}r@{}}Time \\ (s /image) \end{tabular}}} \\
\midrule
\multicolumn{4}{c}{Semi-synthetic data} \\
\midrule
\multirow{2}{*}{\textbf{Resized Image}} & FR & $0.25\pm0.02$ & 2.5 \\ \vspace{0.12cm}
 & NR  & $0.32\pm0.01$ & 1.8 \\
\multirow{2}{*}{\textbf{Patch-based}} & FR & $0.33\pm0.05$ & 20.5 \\
 & NR & $0.45\pm0.04$ & 20.5 \\
\midrule
\multicolumn{4}{c}{Real data} \\
\midrule
\multirow{2}{*}{\textbf{Resized image}} & FR & $0.35 \pm 0.03$ & 1.7 \\ \vspace{0.12cm}
 & NR & $0.43 \pm 0.02$ & 1.7 \\
\multirow{2}{*}{\textbf{Patch-based}} & FR & $0.37 \pm 0.03$ & 17.8\\
 & NR & $0.49 \pm 0.04$ & 20.0\\ 
\bottomrule
\end{tabular}
\end{center}
\end{table}

% Experiment IDs
% FR
%  - 20190927-brj451gvg7-1
%  - 20190926-4155cbh-1
%  - 20190923-4155cbh-1
%  - 20190924-4155cbh-1
% NR
%  - 20190920-4155cbh-1
%  - 20190930-4155cbh-6
%  - 20190930-brj451gvg7-5
%  - 20190930-hp-z840-1
% \begin{table}[b]
% \begin{center}
% \caption{Evaluation on real data data for both inference strategies and full-reference (FR) and no-reference (NR) methods. Values indicate the average and standard deviation for four runs, to account for randomness.}
% \vspace{-0.15cm}
% \label{tab:scores_real_data}
% \begin{tabular}{llrr}
% \toprule
% \textbf{Method} & \textbf{} & \multicolumn{1}{r}{\textbf{mIoU}} &
% \multicolumn{1}{r}{\textbf{\begin{tabular}[c]{@{}r@{}}Time\\ (s /image) \end{tabular}} } \\
% \midrule
% \multirow{2}{*}{\textbf{Resized image}} & FR & $0.35 \pm 0.03$ & 1.7 \\ \vspace{0.12cm}
%  & NR & $0.43 \pm 0.02$ & 1.7 \\
% \multirow{2}{*}{\textbf{Patch-based}} & FR & $0.37 \pm 0.03$ & 17.8\\
%  & NR & $0.49 \pm 0.04$ & 20.0\\
% \bottomrule
% \end{tabular}
% \end{center}
% \end{table}

We evaluated our method on two datasets, one with semi-synthetic defects and another with real samples from defective printers, both on an Intel Xeon E5-2620 CPU and NVIDIA Titan X Maxwell GPU. Due to differences between the two datasets, we trained two types of models for evaluation, one considering banding defects with multiple labels and the other associating banding with a single label.

\begin{figure*}[t]
     \centering
     \begin{subfigure}[b]{0.18\textwidth}
         \centering
         \includegraphics[width=.8\textwidth]{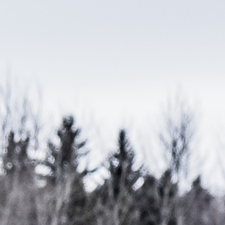}
         \caption{}
         \label{fig:results_semi_real_reference}
     \end{subfigure}
     \begin{subfigure}[b]{0.18\textwidth}
         \centering
         \includegraphics[width=.8\textwidth]{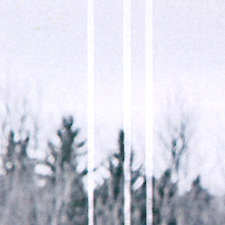}
         \caption{}
         \label{fig:results_semi_real_original}
     \end{subfigure}
     \begin{subfigure}[b]{0.18\textwidth}
         \centering
         \includegraphics[width=.8\textwidth]{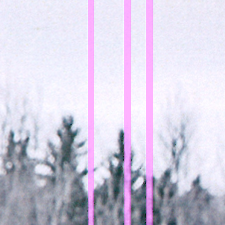}
         \caption{}
         \label{fig:results_semi_real_expected}
     \end{subfigure}
     \begin{subfigure}[b]{0.18\textwidth}
         \centering
         \includegraphics[width=.8\textwidth]{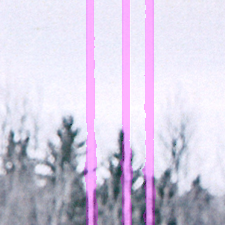}
         \caption{}
         \label{fig:results_semi_real_noref}
     \end{subfigure}
     \begin{subfigure}[b]{0.18\textwidth}
         \centering
         \includegraphics[width=.8\textwidth]{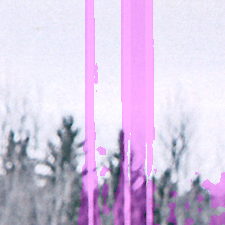}
         \caption{}
         \label{fig:results_semi_real_fullref}
     \end{subfigure}
        %\vspace{-0.1cm}
        \caption{Sample results on the semi-synthetic dataset with patch-based inference mode: (a) reference image input, (b) defective image input, (c) ground truth overlay, predictions from (d) no-reference and (e) full-reference model. Banding defects are marked as magenta. Best viewed on screen.}
        \label{fig:results_semi_real}
\end{figure*}

For the semi-synthetic dataset, we added synthetic artifacts to $103$ background images using the process described in Section~\ref{subsec:artifacts}, printed and scanned them at $600$ dpi (around $7000 \times 5000$ pixels), and aligned the digitized and reference images with feature matching. After alignment, the images had around $2000 \times 3000$ pixels. The printers and scanners used in this dataset were the same used to design our print-scan simulation described in Section~\ref{subsec:artifacts}. Manual annotation was not needed as we recorded defect locations during artifacts generation. Multiple labels were used for banding, depending on the affected color channel (cyan, magenta, yellow, or black), allowing us to examine more detailed predictions for this type of defect. Results for the semi-synthetic dataset are shown in Table~\ref{tab:scores} and Figure~\ref{fig:results_semi_real}.

\begin{figure*}[t]
     \centering
     \begin{subfigure}[b]{0.15\textwidth}
         \includegraphics[width=0.92\textwidth]{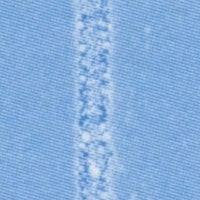}
         \centering
     \end{subfigure}
     \begin{subfigure}[b]{0.15\textwidth}
         \includegraphics[width=0.92\textwidth]{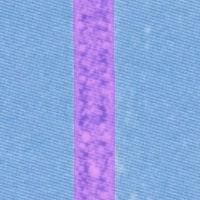}
         \centering
     \end{subfigure}
     \begin{subfigure}[b]{0.15\textwidth}
         \includegraphics[width=0.92\textwidth]{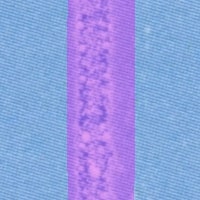}
         \centering
     \end{subfigure}
     \begin{subfigure}[b]{0.15\textwidth}
         \includegraphics[width=0.92\textwidth]{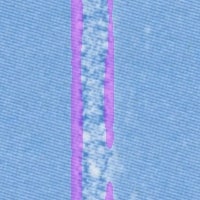}
         \centering
     \end{subfigure}
\par\smallskip
     \begin{subfigure}[b]{0.15\textwidth}
         \includegraphics[width=0.92\textwidth]{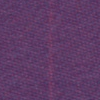}
         \centering
     \end{subfigure}
     \begin{subfigure}[b]{0.15\textwidth}
         \includegraphics[width=0.92\textwidth]{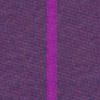}
         \centering
     \end{subfigure}
     \begin{subfigure}[b]{0.15\textwidth}
         \includegraphics[width=0.92\textwidth]{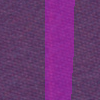}
         \centering
     \end{subfigure}
     \begin{subfigure}[b]{0.15\textwidth}
         \includegraphics[width=0.92\textwidth]{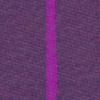}
         \centering
     \end{subfigure}
     \par\smallskip
    \begin{subfigure}[b]{0.15\textwidth}
         \includegraphics[width=0.92\textwidth]{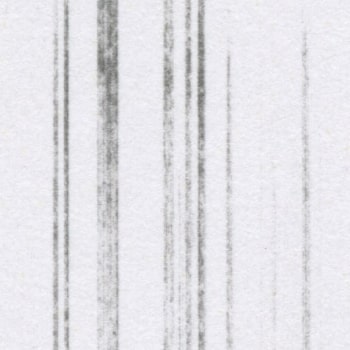}
         \centering
     \end{subfigure}
     \begin{subfigure}[b]{0.15\textwidth}
         \includegraphics[width=0.92\textwidth]{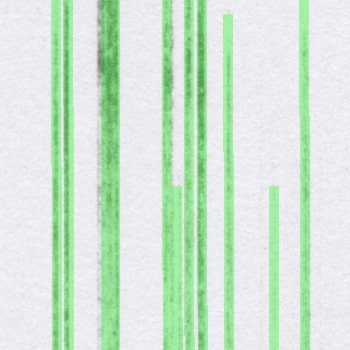}
         \centering
     \end{subfigure}
     \begin{subfigure}[b]{0.15\textwidth}
         \includegraphics[width=0.92\textwidth]{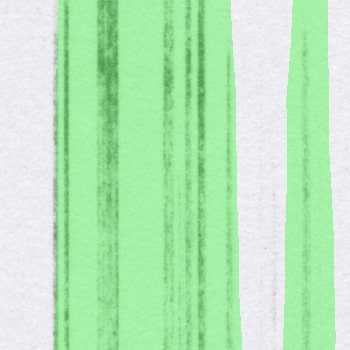}
         \centering
     \end{subfigure}
     \begin{subfigure}[b]{0.15\textwidth}
         \includegraphics[width=0.92\textwidth]{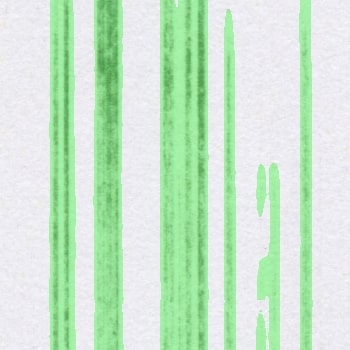}
         \centering
     \end{subfigure}
     \par\smallskip
     \begin{subfigure}[b]{0.15\textwidth}
         \includegraphics[width=0.92\textwidth]{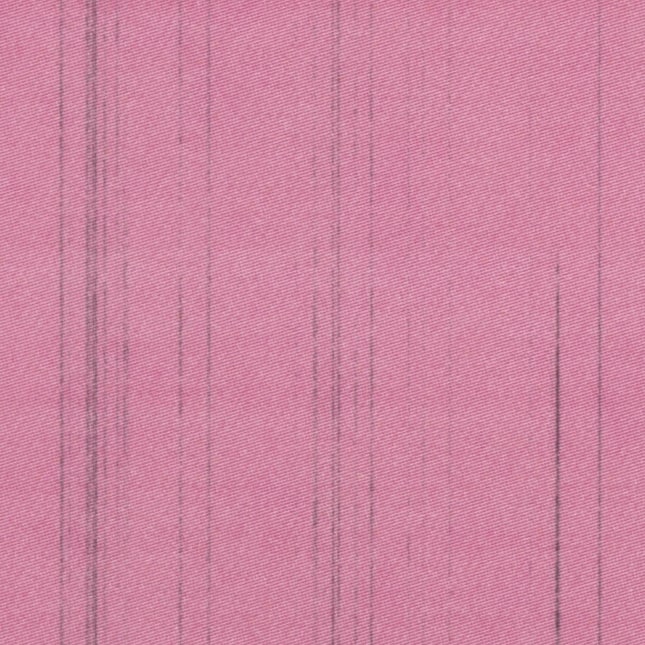}
         \centering
         \caption*{Input}
     \end{subfigure}
     \begin{subfigure}[b]{0.15\textwidth}
         \includegraphics[width=0.92\textwidth]{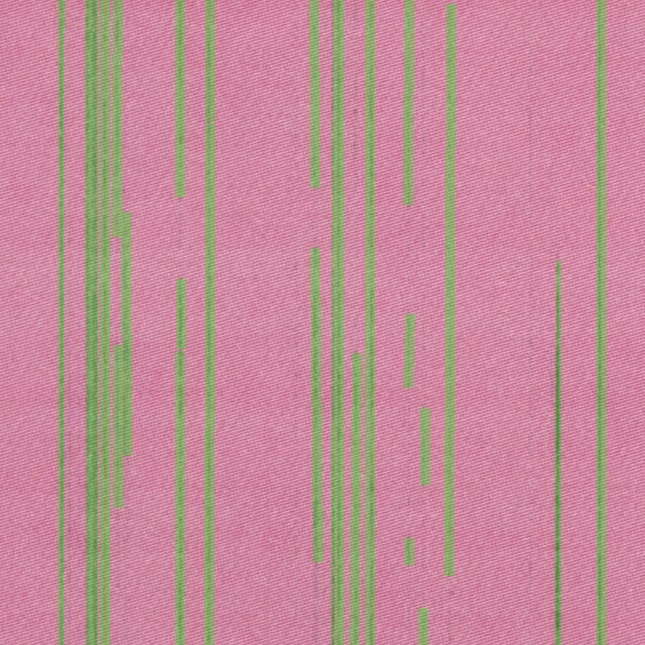}
         \centering
         \caption*{Ground Truth}
     \end{subfigure}
     \begin{subfigure}[b]{0.15\textwidth}
         \includegraphics[width=0.92\textwidth]{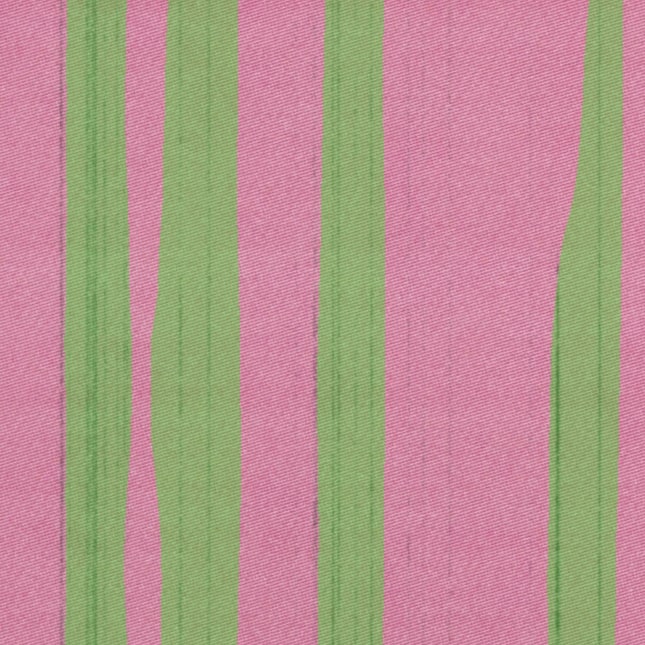}
         \centering
         \caption*{Resized Image}
     \end{subfigure}
     \begin{subfigure}[b]{0.15\textwidth}
         \includegraphics[width=0.92\textwidth]{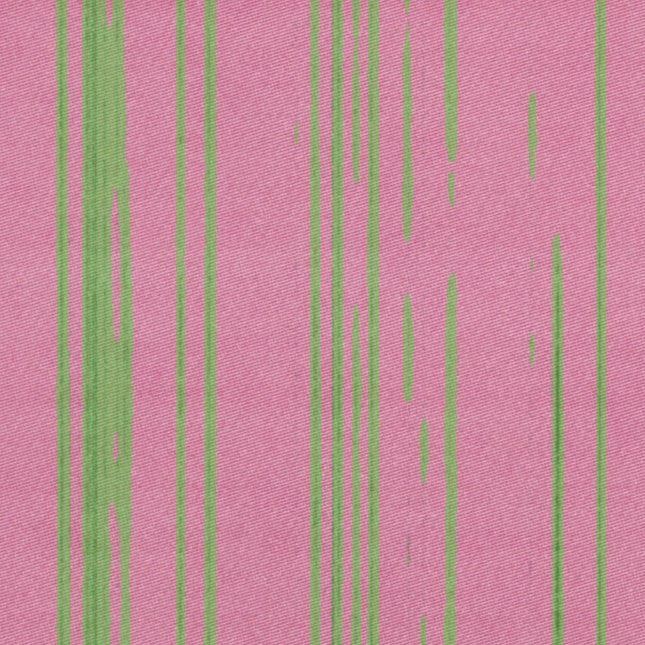}
         \centering
         \caption*{Patch-based}
     \end{subfigure}
        %\vspace{-0.1cm}
        \caption{Sample results on the real dataset for the no-reference method with resized image and patch-based inference strategies. Streaks are marked as green and banding defects are marked as magenta. Note that patch-based inference produces finer results than resized image. Best viewed on screen.}
        \label{fig:compare}
\end{figure*}

The real dataset served as a benchmark of how well the synthetic training data was translating to the real-world. It has $30$ scanned images with real banding and streaks defects, collected from defective printers, and scanned at $600$ dpi. We manually annotated the defects on each scanned image and manually aligned the digitized and reference images. After alignment, the images had average dimensions of $2500 \times 3000$. A single label was assigned to banding defects, due to the difficulty of manually analyzing in retrospect which color channel had been affected. Table~\ref{tab:scores} and Figure~\ref{fig:compare} show the results for the real dataset.

% COMMANDS TO GET CROPS
% ```
% convert 173_a.png -crop 225x225+1370+110 semireal-banding-example-reference.png
% convert 173_b.png -crop 225x225+1370+110 semireal-banding-example-original.png
% convert 173_c.png -crop 225x225+1370+110 semireal-banding-example-expected.png
% convert 173_d.png -crop 225x225+1370+110 semireal-banding-example-noreference.png
% convert 173_e.png -crop 225x225+1370+110 semireal-banding-example-fullreference.png
% ```

% COMMANDS TO GET CROPS
% ```
% convert 14_gt.png -crop 200x200+3040+2350 14_gt.png
% convert 14_orig.png -crop 200x200+3040+2350 14_orig.png
% convert 14_resize.png -crop 200x200+3040+2350 14_resize.png
% convert 14_slide.png -crop 200x200+3040+2350 14_slide.png
% convert 27_gt.png -crop 350x350+2050+1450 27_gt.png
% convert 27_orig.png -crop 350x350+2050+1450 27_orig.png
% convert 27_resize.png -crop 350x350+2050+1450 27_resize.png
% convert 27_slide.png -crop 350x350+2050+1450 27_slide.png
% convert 32_gt.png -crop 645x645+1935+2350 32_gt.crop.png
% convert 32_gt.png -crop 645x645+1935+2350 32_gt.png
% convert 32_orig.png -crop 645x645+1935+2350 32_orig.png
% convert 32_resize.png -crop 645x645+1935+2350 32_resize.png
% convert 32_slide.png -crop 645x645+1935+2350 32_slide.png
% convert 41_gt.png -crop 100x100+1260+5500 41_gt.png
% convert 41_orig.png -crop 100x100+1260+5500 41_orig.png
% convert 41_resize.png -crop 100x100+1260+5500 41_resize.png
% convert 41_slide.png -crop 100x100+1260+5500 41_slide.png
% ```

Following the semantic segmentation literature, we evaluated our models with the mean intersection over union (mIoU) over classes. Previous works on print defect detection~\cite{wang2016, runzhe2019, xiao2017} do not evaluate their approaches in terms of mIoU, and each of them is evaluated differently, making direct comparisons among works hard.

\section{Discussion}\label{sec:discussion}

Table~\ref{tab:scores} shows that the NR method outperforms FR on all datasets and inference approaches. At first glance, this result seems surprising as the reference image could provide cues for the model. Nevertheless, we hypothesize that it may be difficult for the FR network to realize that it should take into account the reference and scanned images separately when predicting the defective pixels since concatenating them does not explicitly convey that.

The human eye is capable of detecting defects without a reference image. Hopefully, a machine learning model may be able to do the same. State of the art in computer vision shows that deep learning methods usually work well with minimal supervision than other methods, which may strongly rely on handcraft features. It may be the case that the NR method works better with DCNNs because the input space (3 channels) is simpler than in the FR method (6 channels), facilitating convergence.

Figures~\ref{fig:results_semi_real} and~\ref{fig:compare} show that the patch-based inference provided more fine-grained detection compared to the resized image mode. However, this advantage comes with the cost of it being considerably slower (Table~\ref{tab:scores}). Choosing between one of the inference methods depends on the goal of the problem. If the system must be fast, resized image is the best choice, while patch-based inference is favored when quality is more important.

Previous works~\cite{wang2016, runzhe2019} require input images to have specific aspects such as the ones of standard print test pages, which are mostly solid-color filled. Conversely, our framework is more general, working for documents, photos, and combinations of both. Furthermore, our framework does not require calibrating thresholds for different occurrences of streak defects, as in~\cite{runzhe2019}.

\begin{figure}[ht]
  \centering
    \begin{subfigure}{.22\linewidth}
        \centering
        \includegraphics[width=\textwidth]{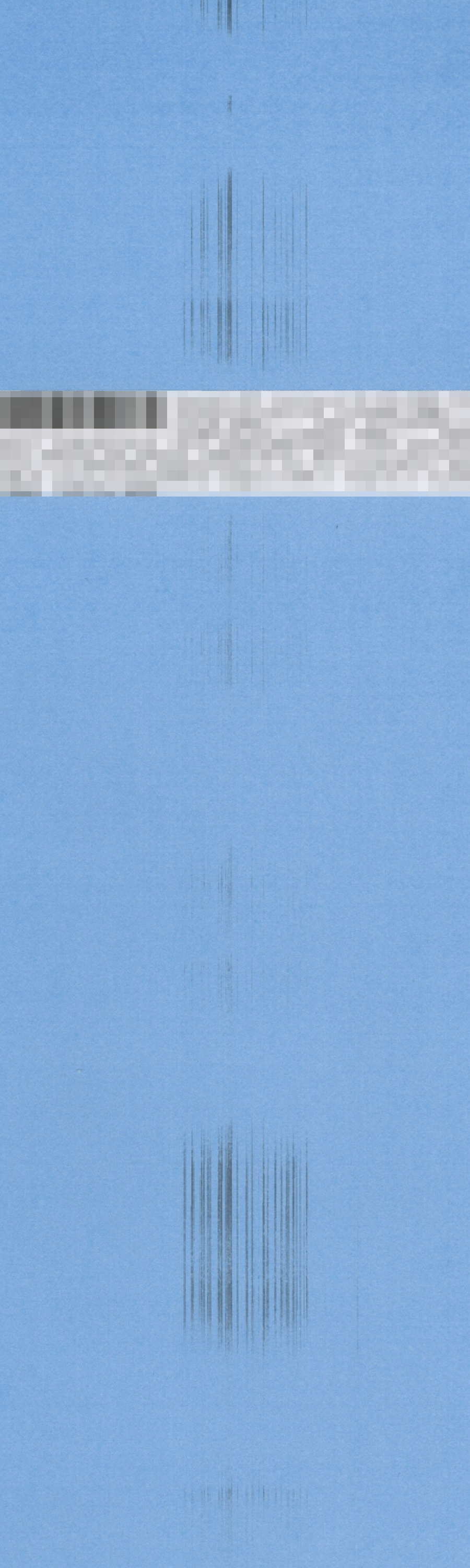}
        \caption{}
    \end{subfigure}
    \begin{subfigure}{.22\linewidth}
        \centering
        \includegraphics[width=\textwidth]{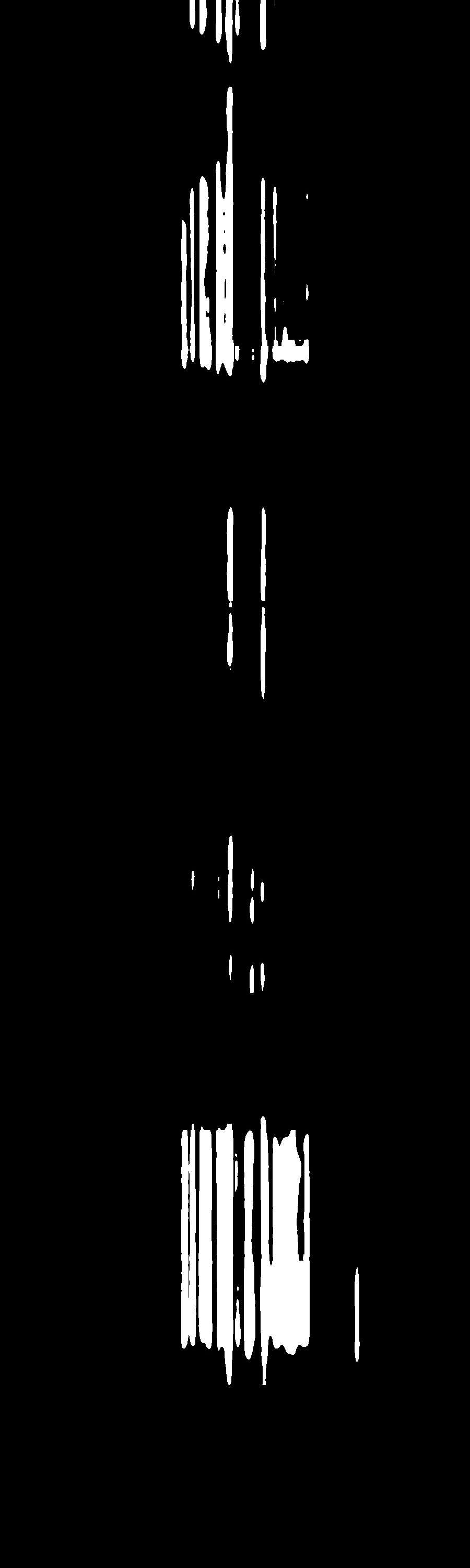}
        \caption{}
    \end{subfigure}
    \begin{subfigure}{.22\linewidth}
        \centering
        \includegraphics[width=\textwidth]{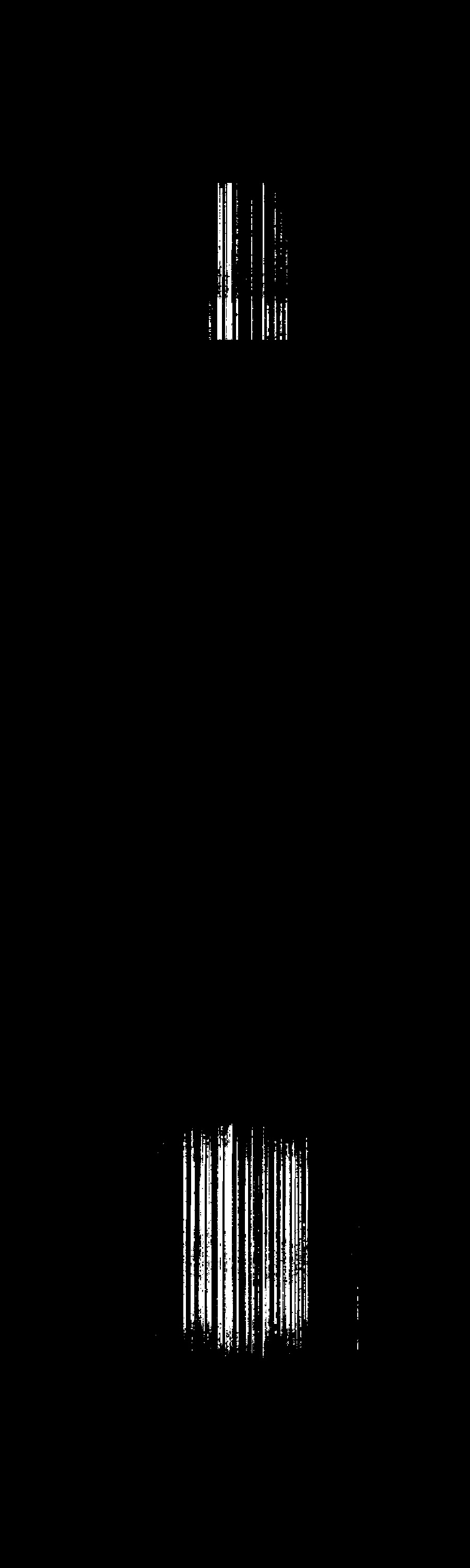}
        \caption{}
    \end{subfigure}
    \caption{Comparison between methods: (a) real image crop with streaks (b) map of our method (c) method by \cite{wang2016}. Best viewed on screen.}
    \label{fig:method_comparison}
\end{figure}

Considering the lack of similar works, we found it suitable to use an intermediate step in the framework proposed by Wang \etal~\cite{wang2016} as a base for comparison. Their method provides a fine defect detection map from which we could compute the IoU and compare it with ours. Due to input restrictions in their framework, we evaluated it only with solid-color filled defective images, as depicted in Figure~\ref{fig:method_comparison}. 

We chose \textit{patch-based} inference as it provides better detection of streaks, and the NR method to agree with Wang \etal's detection method, which does not require the reference image. Our model was able to detect more streaks, even the fading ones, when compared to their approach. This can be verified by a higher IoU value of 0.299, outperforming their mapping, which achieved an IoU of 0.183. Despite that, we would like to emphasize that Wang \etal's aim was different from ours. Their defect detection was an intermediate step to locate and get the defect features to subsequently feed a predictor and get an overall quality assessment of the printed image.

Although we use the mIoU to evaluate the performance of our models, we realized from our experience annotating the evaluation dataset that two human annotators might produce different segmentation masks, introducing biases towards annotation. This problem is recurrent in other applications of semantic segmentation~\cite{ribeiro2019handling}. In print defect mapping, since defects may be very thin, annotation agreement may even be smaller. Thus, for more realistic evaluation, we may use some metric relaxation in future works.

\section{Conclusion and Future Works}

We presented an end-to-end technique that uses DCNNs to perform semantic segmentation for print defect mapping. To the best of our knowledge, this is the first work to describe a fully-automated pipeline for detecting print defects at the pixel level.

We simulated printing and scanning of documents, thus eliminating the need for this laborious task. Moreover, our method continuously simulates data, making our pipeline cost-free in terms of annotation. Our results show that our models can transfer the knowledge learned from synthetic data to real print defects, demonstrating its applicability for print quality assessment in industry.

We can improve our framework by updating specific parts of the pipeline. The DCNN can be swapped to a more modern architecture. Models may also take advantage of attention mechanisms to improve the results by making the network focus more on specific areas of the image. The preprocessing pipeline can be further improved with better defects simulations and with stronger data augmentation. Additionally, performance can increase with larger datasets.

The framework can be improved by including more types of defects so that a single approach can be used to detect a wide range of print defects. Nevertheless, our results show that it is possible to have two defects in the same model, indicating that we may also get good results when adding more defect types. Since we use synthetic defects, the hard part would be to simulate the defects.

Alongside with the defect mapping, the framework could also predict a global quality score in a multi-task learning approach. A global quality score can be useful to quantify the overall condition of a printer and can be used to discard defective pages automatically. However, we need to define some heuristics in order to produce synthetic scores for the training set.

\ifwacvfinal
\section*{Acknowledgements}
We thank Eric Maggard from HP Inc. at Boise, ID for providing the image dataset with real print defects, and Jianyu Wang, Terry Nelson, Renee Jessome, Steve Astling, Eric Maggard, Mark Shaw, and Jan P. Allebach for providing us the source code from~\cite{wang2016}.

\else
\fi

{\small
\bibliographystyle{ieee}
\bibliography{egbib}

\begin{thebibliography}{10}\itemsep=-1pt

\bibitem{berman2018lovasz}
M.~Berman, A.~Rannen~Triki, and M.~B. Blaschko.
\newblock The lov{\'a}sz-softmax loss: a tractable surrogate for the
  optimization of the intersection-over-union measure in neural networks.
\newblock In {\em Proceedings of the IEEE Conference on Computer Vision and
  Pattern Recognition}, pages 4413--4421, 2018.

\bibitem{bosse2018}
S.~{Bosse}, D.~{Maniry}, K.~{Müller}, T.~{Wiegand}, and W.~{Samek}.
\newblock Deep neural networks for no-reference and full-reference image
  quality assessment.
\newblock {\em IEEE Transactions on Image Processing}, 27(1):206--219, Jan
  2018.

\bibitem{briggs2000}
J.~C. Briggs.
\newblock Banding characterization for inkjet printing.
\newblock In {\em 2000 Image Processing, Image Quality, Image Capture Systems
  Conference}, March 2000.

\bibitem{chen2018deeplabv2}
L.~Chen, G.~Papandreou, I.~Kokkinos, K.~Murphy, and A.~L. Yuille.
\newblock Deeplab: Semantic image segmentation with deep convolutional nets,
  atrous convolution, and fully connected crfs.
\newblock {\em IEEE Transactions on Pattern Analysis and Machine Intelligence},
  40(4):834--848, April 2018.

\bibitem{chen2018deeplabv3plus}
L.-C. Chen, Y.~Zhu, G.~Papandreou, F.~Schroff, and H.~Adam.
\newblock Encoder-decoder with atrous separable convolution for semantic image
  segmentation.
\newblock In V.~Ferrari, M.~Hebert, C.~Sminchisescu, and Y.~Weiss, editors,
  {\em Computer Vision -- ECCV 2018}, pages 833--851, Cham, 2018. Springer
  International Publishing.

\bibitem{jang2005simulation}
W.~Jang and J.~P. Allebach.
\newblock Simulation of print quality defects.
\newblock {\em Journal of Imaging Science and Technology}, 49(1):1--18, 2005.

\bibitem{kang2014}
L.~{Kang}, P.~{Ye}, Y.~{Li}, and D.~{Doermann}.
\newblock Convolutional neural networks for no-reference image quality
  assessment.
\newblock In {\em 2014 IEEE Conference on Computer Vision and Pattern
  Recognition}, pages 1733--1740, June 2014.

\bibitem{kim2017fully_deep_blind}
J.~Kim and S.~Lee.
\newblock Fully deep blind image quality predictor.
\newblock {\em IEEE Journal of Selected Topics in Signal Processing},
  11(1):206--220, Feb 2017.

\bibitem{KingmaB14}
D.~P. Kingma and J.~Ba.
\newblock Adam: {A} method for stochastic optimization.
\newblock {\em CoRR}, abs/1412.6980, 2014.

\bibitem{li2016_nr_iqa_dnn}
Y.~Li, L.~Po, L.~Feng, and F.~Yuan.
\newblock No-reference image quality assessment with deep convolutional neural
  networks.
\newblock In {\em 2016 IEEE International Conference on Digital Signal
  Processing (DSP)}, pages 685--689, Oct 2016.

\bibitem{long2015fcns}
J.~Long, E.~Shelhamer, and T.~Darrell.
\newblock Fully convolutional networks for semantic segmentation.
\newblock In {\em 2015 IEEE Conference on Computer Vision and Pattern
  Recognition (CVPR)}, pages 3431--3440, June 2015.

\bibitem{pan2018blind_predicting}
D.~Pan, P.~Shi, M.~Hou, Z.~Ying, S.~Fu, and Y.~Zhang.
\newblock Blind predicting similar quality map for image quality assessment.
\newblock In {\em 2018 IEEE/CVF Conference on Computer Vision and Pattern
  Recognition}, pages 6373--6382, June 2018.

\bibitem{ribeiro2019handling}
V.~Ribeiro, S.~Avila, and E.~Valle.
\newblock Handling inter-annotator agreement for automated skin lesion
  segmentation.
\newblock {\em arXiv preprint arXiv:1906.02415}, 2019.

\bibitem{ronneberger2015unet}
O.~Ronneberger, P.~Fischer, and T.~Brox.
\newblock U-net: Convolutional networks for biomedical image segmentation.
\newblock In N.~Navab, J.~Hornegger, W.~M. Wells, and A.~F. Frangi, editors,
  {\em Medical Image Computing and Computer-Assisted Intervention -- MICCAI
  2015}, pages 234--241, Cham, 2015. Springer International Publishing.

\bibitem{siam2018rtseg}
M.~Siam, M.~Gamal, M.~Abdel-Razek, S.~Yogamani, and M.~Jagersand.
\newblock Rtseg: Real-time semantic segmentation comparative study.
\newblock In {\em 2018 25th IEEE International Conference on Image Processing
  (ICIP)}, pages 1603--1607, Oct 2018.

\bibitem{spivakovsky2018image}
A.~Spivakovsky, O.~Perry, O.~Haik, and A.~Malki.
\newblock Image defect detection, May~10 2018.
\newblock US Patent US20180131815A1.

\bibitem{Verikas:2011:RAC:1994471.1994864}
A.~Verikas, J.~Lundstr\"{o}m, M.~Bacauskiene, and A.~Gelzinis.
\newblock Review: Advances in computational intelligence-based print quality
  assessment and control in offset colour printing.
\newblock {\em Expert Syst. Appl.}, 38(10):13441--13447, Sept. 2011.

\bibitem{wang2016}
J.~Wang, T.~Nelson, R.~Jessome, S.~Astling, E.~Maggard, M.~Shaw, and J.~P.
  Allebach.
\newblock Local defect detection and print quality assessment.
\newblock {\em Electronic Imaging}, 2016(13):1--7, 2016.

\bibitem{xiao2017}
Z.~Xiao, M.~Nguyen, E.~Maggard, M.~Shaw, J.~Allebach, and A.~Reibman.
\newblock Real-time print quality diagnostics.
\newblock {\em Electronic Imaging}, 2017(12):174--179, 2017.

\bibitem{yangping2018real}
W.~Yangping, X.~Shaowei, Z.~Zhengping, S.~Yue, and Z.~Zhenghai.
\newblock Real-time defect detection method for printed images based on
  grayscale and gradient differences.
\newblock {\em Journal of Engineering Science \& Technology Review}, 11(1),
  2018.

\bibitem{runzhe2019}
R.~Zhang, E.~Maggard, R.~Jessome, M.~Cho, and J.~P. Allebach.
\newblock Block window method with logistic regression algorithm for streak
  detection.
\newblock {\em Electronic Imaging}, 2019:300--1--300--7, 2019.

\bibitem{remote_sensing_xx_zhu}
X.~X. {Zhu}, D.~{Tuia}, L.~{Mou}, G.~{Xia}, L.~{Zhang}, F.~{Xu}, and
  F.~{Fraundorfer}.
\newblock Deep learning in remote sensing: A comprehensive review and list of
  resources.
\newblock {\em IEEE Geoscience and Remote Sensing Magazine}, 5(4):8--36, Dec
  2017.

\bibitem{zlateski2018importance}
A.~Zlateski, R.~Jaroensri, P.~Sharma, and F.~Durand.
\newblock On the importance of label quality for semantic segmentation.
\newblock In {\em Proceedings of the IEEE Conference on Computer Vision and
  Pattern Recognition}, pages 1479--1487, 2018.

\end{thebibliography}
}

\end{document}